\documentclass[10pt,letterpaper]{article}
\usepackage[a4paper,
            bindingoffset=0.2in,
            left=1in,
            right=1in,
            top=1in,
            bottom=1in,
            footskip=.25in
            ]{geometry}
\usepackage[normalem]{ulem}
\useunder{\uline}{\ul}{}
\usepackage{multirow}
\usepackage{graphicx}
\usepackage{times}
\usepackage{float} 
\usepackage{subfigure} 
\usepackage{tabularx,booktabs}
\usepackage{url}
\usepackage{fullpage}
\usepackage{graphicx}
\usepackage{lipsum}
\newcommand{\junk}[1]{}

\usepackage{helvet}
\usepackage{courier}
\usepackage{blindtext}
\usepackage{enumitem}
\usepackage{xcolor}
\usepackage{lineno}
\usepackage{caption}

\usepackage{xr}
\makeatletter

\usepackage{amsmath}
\usepackage{indentfirst}
\usepackage{authblk}
\usepackage{amsfonts} 
\usepackage{relsize}
\begin{document}
\thispagestyle{empty}

\title{Unbiased organism-agnostic and highly sensitive signal peptide predictor with deep protein language model}

\author[1,4]{Junbo Shen\thanks{Equal first authorship.}}
\author[1]{Qinze Yu$^\ast$}
\author[1,2,5]{Shenyang Chen$^\ast$}
\author[1]{Qingxiong Tan}
\author[1]{Jingchen Li}
\author[1,2,3,6,7,8]{Yu Li \thanks{Corresponding Author. Email: liyu@cse.cuhk.edu.hk}}
\affil[1]{\small Department of Computer Science and Engineering, CUHK, Hong Kong SAR, China}
\affil[2]{\small The CUHK Shenzhen Research Institute, Hi-Tech Park, Nanshan, Shenzhen, 518057, China}
\affil[3]{\small Shanghai Artificial Intelligence Laboratory, Shanghai, China}
\affil[4]{\small Department of Computer Science and Engineering, Washington University, St. Louis, MO 63130, United States}
\affil[5]{\small Georgia Institute of Technology, Atlanta, GA 30332, United States}
\affil[6]{Institute for Medical Engineering and Science, Massachusetts Institute of Technology, Cambridge, MA, USA}
\affil[7]{Wyss Institute for Biologically Inspired Engineering, Harvard University, Boston, MA, USA}
\affil[8]{Broad Institute of MIT and Harvard, Cambridge, MA, USA}

\date{}
\maketitle
\begin{abstract}

Signal peptide (SP) is a short peptide located in the N-terminus of proteins. It is essential to target and transfer transmembrane and secreted proteins to correct positions. Compared with traditional experimental methods to identify signal peptides, computational methods are faster and more efficient, which are more practical for analyzing thousands or even millions of protein sequences, especially for metagenomic data. Computational tools are recently proposed to classify signal peptides and predict cleavage site positions. However, most of them disregard the extreme data imbalance problem in these tasks. 
In addition, almost all these methods rely on additional group information of proteins to boost their performances, which, however, may not always be available.
To address these issues, we present Unbiased Organism-agnostic Signal Peptide Network (USPNet), a signal peptide classification and cleavage site prediction deep learning method that takes advantage of protein language models. We propose to apply label distribution-aware margin loss to handle data imbalance problems and use evolutionary information of protein to enrich representation and overcome species information dependence. Extensive experimental results demonstrate that our proposed method significantly outperforms all the previous methods on classification performance by 10\% on multiple criteria. Additional studies on the simulated proteome data and organism-agnostic experiments further indicate that our model is a more universal and robust tool without dependency on additional group information of proteins. Building on this, we design a whole signal-peptides-discovering pipeline to explore unprecedented signal peptides from metagenomic data. The proposed method is highly sensitive and reveals 347 predictions to be the candidate novel SPs, with the lowest sequence identity between our candidate peptides and the closest signal peptide in the training dataset at only 13\%. Interestingly, the TM-scores between candidates and SPs in the training set are mostly above 0.8. The further analysis of the experimentally verified novel SPs provides evidence that although USPNet does not rely on any structure information as input, it detects SPs based on evolutionary and structural information instead of sequence-similarity. The results showcase that USPNet has learned SP structure with just raw amino acid sequences and the large protein language model, and thus enables the discovery of novel SPs that are distant from existing knowledge effectively.



\end{abstract}
\newpage

\section{Introduction}

A signal peptide (SP) is a short amino acid sequence working as a specific targeting signal to guide and transfer proteins into secretory pathways \cite{von1998life}. It has a three-domain structure: Positively charged N-region, hydrophobic H-region, and uncharged C-region \cite{r1}. The SPs function as specific segments to guide proteins to reach correct positions and then be cleaved by cleavage sites nearby their C region. Thus, the identification of signal peptides is vital for studying destinations and functions of proteins \cite{bradshaw2009signal,craig2019type, duan2018signal, jiang2022n}. 

As comprehensive experimental identification of SPs can be time- and resources-consuming, many computational tools have been proposed to classify signal peptides and predict cleavage sites. The first attempt was a formulating rule proposed in 1983 \cite{r3}. Von Heijne first applied a statistical method to unveil patterns nearby cleavage sites of signal peptides based on only 78 Eukaryotic proteins \cite{r3}. Furthermore, generative models, such as the hidden Markov model (HMM), were proposed to facilitate the recognition of signal peptides. These models focused on analyzing these three functional regions (N-region, C-region, and H-region) in detail and were built by capturing the relationships between different regions of signal peptide \cite{r7,r8,ehsan2018novel, janda2010recognition, madani2023large}.
Different from generative models, some homology-based methods were proposed \cite{r11}. The predictions of these methods are based on the similarities between sequences in the existing knowledge base and the input sequences. In addition, they can achieve similar prediction performance as generative models. 

Recently, supervised models have made great progress in signal peptide recognition. The query sequences are encoded into embedding vectors and then fed into models to directly compute probabilities for each signal peptide type. Among these methods, machine-learning-based models play an important role in their remarkable performances \cite{petersen2011signalp}. DeepSig applied deep convolutional neural networks (DCNNs) architecture to the recognition of signal peptides and the prediction of cleavage site positions \cite{r2}.  Furthermore, SignalP5.0 came up and benchmarked all the previously proposed methods \cite{r14}, and SignalP6.0 \cite{signalp6.0} is able to predict all 5 types of signal peptides that the previous model failed to detect. These methods achieved advanced performance in tasks, but most of them suffered from extreme class imbalance and therefore performed poorly on data from minor classes \cite{lipop, PRED-LIPO, Tatp}.
In addition, these methods often depend heavily on additional information about groups of organisms to boost their performances. However, it is impractical to obtain sufficient group information from metagenomic data in reality \cite{pasolli2017accessible, sczyrba2017critical}. A robust tool should only require amino acid sequences to yield accurate prediction results. 

In signal peptide classification, the key problems to solve are the imbalance of training data and object dependence on group information. 
Rao et al. \cite{r16} introduced a transformer-based language model, ESM-1b, which demonstrated that information learned from large-scale protein sequences alone could implicitly encode functional and structural information and benefits various downstream tasks, such as secondary structure prediction and contact prediction, outperforming specific-data-trained model by a large margin. Rao et al. \cite{MSA} also found that integrating multiple sequence alignments (MSA) into the model, referred to as the MSA transformer, led to more excellent performance. These protein language models \cite{biswas2021low, alley2019unified} gain improvement on problems with limited annotated data. Inspired by that, 
we thus propose the unbiased organism-agnostic signal peptide predictor (USPNet) based on a BiLSTM \cite{bi} framework and protein language models to classify signal peptides and predict their cleavage site positions. We leverage an advanced MSA-based protein language model to enrich the representations to aid in encoding group information of sequences. Moreover, we combine class-balance loss with Label distribution-aware margin (LDAM) loss \cite{r15} as the loss function of USPNet to improve generalization.
Our model is end-to-end, which takes just raw amino acids as inputs. It is efficient to perform the classification of all five types of signal peptides and the non-signal-peptide type protein. We compare our model with several task-related deep learning models on the re-classified SignalP5.0 benchmark set.
Notably, USPNet achieves more than 10\% improvement of MCC on multiple classes compared to the previous state-of-the-art methods. Besides, our model significantly outperforms SignalP6.0 on the recall rate of cleavage site prediction. On our curated domain-shift independent set, USPNet also performs better than other models, showing the generalization of our method on the classification of signal peptides. To further showcase the potency of USPNet, we collect proteome-wide data from Escherichia coli (strain K12) as well as other 7 organisms. When applying USPNet to detect signal peptides from proteome-wide data, it retrieves nearly all of them, which is among the best of all models being tested. In addition, we thoroughly assessed the group information dependency by conducting an organism-agnostic experiment which removes group information in the input. USPNet's well-trained encoders empowered with MSA information as well as protein language models capture rich evolution and functional information,  allowing the model to remain robust despite the absence of group information.
Such highly sensitive signal peptide prediction capability enables novel SPs mining from large metagenomic resources. We thus build a complete pipeline from handling metagenomic data to making novel signal peptide detection. We collect swine gut metagenomics data from multiple resources to carry out the case study and finally screen out 347 peptides from millions of sequences as the candidates that have low sequence identity with existing SPs and are likely to be novel signal peptides. The result indicates that USPNet is able to provide evolutionary and structural information of SPs and effectively discovered candidate signal peptides that are distant from existing knowledge.

%


The main contributions of our work are summarized as follows:
\begin{itemize}

\item[$\bullet$] We introduce Unbiased Organism-agnostic Signal Peptide Network (USPNet), which is able to predict all 5 known types of signal peptides. Extensive experiments demonstrate that the proposed method achieves state-of-the-art performance over other signal peptide predictors on signal peptide classification. We apply USPNet to the independent set and proteome-wide studies. The model reaches 10\% improvement on multiple criteria compared with previous methods and keeps its performance above 90\%.
 
\item[$\bullet$] We provide two versions of USPNet for usage. One constructs multi-sequence alignment (MSA) and uses an MSA transformer to generate embeddings to enrich our representations. 
The other utilizes evolutionary scale modeling (ESM) embeddings \cite{r16}, which we named USPNet-fast. The first version of USPNet has better prediction ability, and USPNet-fast can make inferences 20 times faster. We facilitate users so that they can choose the tool based on the application scenarios.

\item[$\bullet$] We resolve the extreme imbalance problem in signal peptide prediction. Considering that previous algorithms train models mainly based on cross-entropy loss, we propose to apply label-distribution-aware-margin loss (LDAM) to improve the generalization of less frequent classes \cite{r15}. We present a modified loss function by combining class-balance loss with LDAM loss to make USPNet learn useful information from small classes. 

\item[$\bullet$] We build a whole pipeline to detect signal peptides from original metagenomics data. We reveal 347 predictions to be the candidate novel SPs, with the lowest sequence identity between our candidate peptides and the closest signal peptide in the training dataset at only 13\%. Notably, the TM-scores between candidates and SPs in the training set are mostly above 0.8. We also retrieve all 4 experimentally verified signal peptides of our study genomes supported by literature and do not exist in the training dataset. Results show that USPNet learns evolutionary and structural information without additional inputs from the protein folding model and hence could discover peptides that have low sequence identity but high structural similarity at an ideal speed.


\end{itemize}

\section{Result}
\subsection{USPNet is a pipeline for predicting signal peptides from metagenomics data}

\begin{figure}[H] 
\centering 
\includegraphics[width=1.0\textwidth]{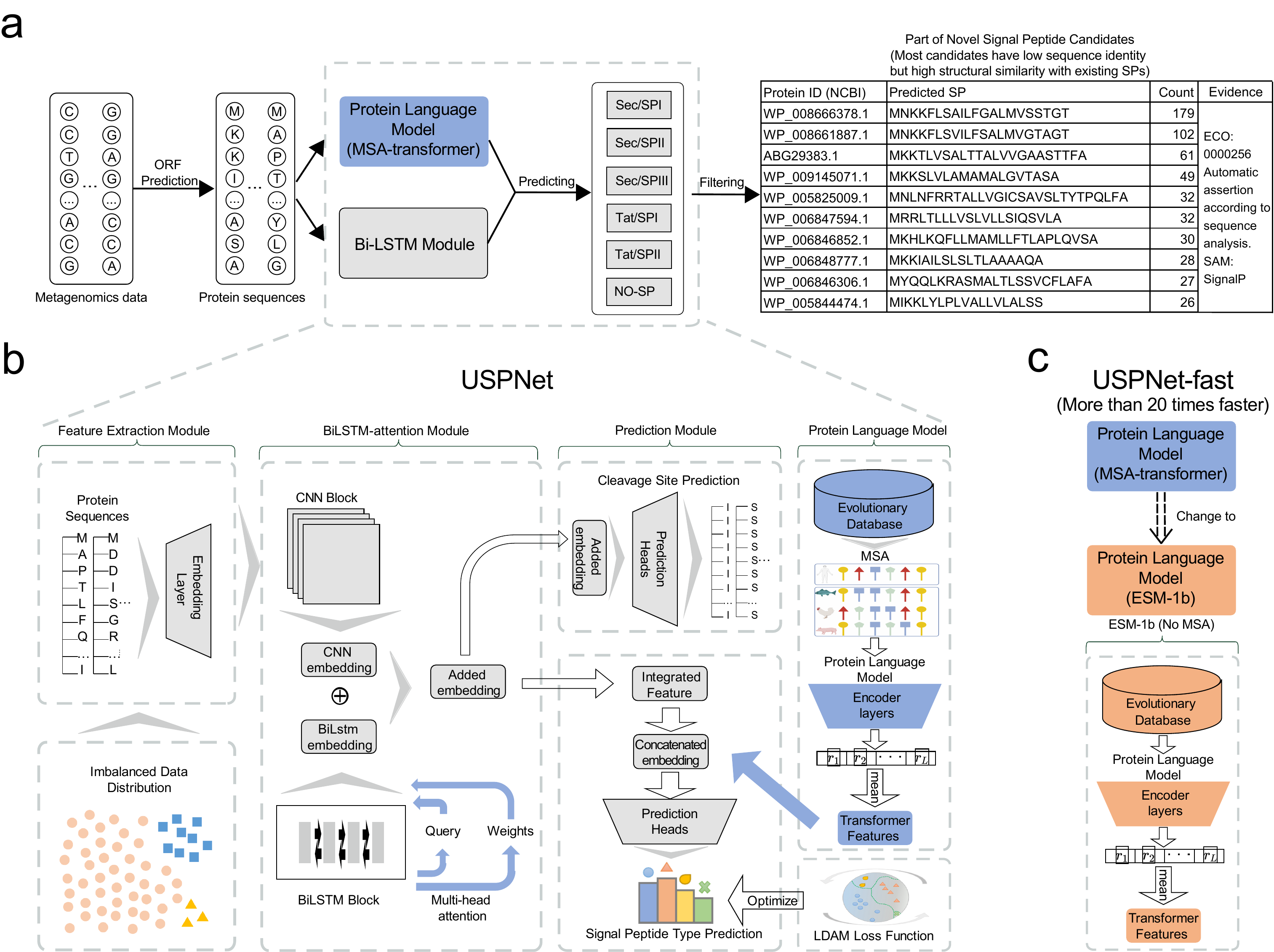}
\caption{\textbf{USPNet workflow for predicting signal peptide (SP) and cleavage site.} \textbf{a}. The pipeline of USPNet for signal peptide discovery from metagenomics data. USPNet takes raw amino acid sequences as input and uses BiLSTM module as well as Protein Language Model to get concatenated embeddings as input for prediction. Here we list the top 10 most frequently occurring peptides that are automatically annotated as SPs by the UniProt database out of 347 candidates. \textbf{b}. Detailed architecture of USPNet. The training data is imbalanced. The protein sequences go through the feature extraction module and are then passed to the BiLSTM module, which includes a Bi-LSTM layer with self-attention and a CNN for extracting long-distance dependencies and features of the sequences.  For SP type prediction, USPNet incorporates MSA embeddings generated by a pre-trained MSA Transformer model. Subsequent MLP-based modules predict cleavage sites and signal peptide types. Label Distribution-Aware Margin (LDAM) loss is employed in training to address data imbalance. \textbf{c}. USPNet-fast replaces the MSA-transformer with ESM-1b, which does not require MSA, and thus enables a much faster inference speed.
} 
\label{main} 
\end{figure}

As shown in Figure \ref{main}.a, the pipeline can predict signal peptides from metagenomics data, and even discover novel SP candidates. The basic architecture of our method is the Bi-LSTM \cite{bi, r17} with the self-attention mechanism \cite{mnih2014recurrent}, and we leverage protein language model-based encoder to enrich representations (Figure \ref{main}.b). USPNet takes amino acid sequences as input and simultaneously predicts signal peptide types and the corresponding cleavage sites. Considering the length of signal peptides in proteins N-terminus is usually between 5 and 30, we set 70 as the cut-off to the length of proteins, which means that each input sequence contains at most 70 amino acids. The sequence then goes to the feature extraction module, employed with an embedding layer. As the number of common residue types is 20, it converts the input sequence into an L×20-dimensional matrix, where L is the length of the sequence. Particularly, we add an L×4-dimensional vector at the head of the generated embedding to store the group information (Eukaryotes, Gram-positive, Gram-negative bacteria, and Archaea). Then, the generated embedding is fed into our BiLSTM part. It consists of a Bi-LSTM layer with self-attention and a CNN, which simultaneously extracts forward and backward directions of long-distance dependencies and global/local features of sequences. Following the Bi-LSTM module, we develop an MLP-based module to predict cleavage sites and SP types separately. To integrate more information in classifying signal peptides, we include MSA embeddings. To be specific, we first generate Multi-sequence alignment (MSA) for each sequence and then input MSAs to the pre-trained MSA Transformer model \cite{MSA} to obtain embeddings from its final layer. And to bend over backward to resolve the data imbalance, our loss function is designed by combining class-balance loss with Label distribution-aware margin (LDAM) loss \cite{r15}. Detailed information is introduced in the Methods part.

Besides the above-mentioned method, we also introduce another version of USPNet, called USPNet-fast, which replaces MSA embeddings with ESM-1b embeddings \cite{r16}, and keeps other modules unchanged (Figure \ref{main}.c). USPNet-fast is able to predict signal peptide and cleavage sites faster and without much performance degeneration. 

\subsection{USPNet outperforms the previous methods on the benchmark dataset}
\subsubsection*{Signal peptide type prediction}

USPNet is able to predict the type and cleavage site of a signal peptide at the same time. To fair analyze performance, we train and test our model on the re-classified, extended, and homology-reduced datasets derived from the data published with SignalP5.0 and SignalP6.0\cite{r14,signalp6.0}. The combination of training and benchmark data is identical to the homology partitioned SignalP6.0 dataset. For signal peptide type prediction, the training data, with benchmark data excluded from the dataset, contains 13679 sequences and can be considered into six parts according to six different kinds of labels. However, the data is extremely imbalanced. As shown in Table \ref{DSTotal}, the sequences with major-class labels are ten times more than those with minor-class labels. Therefore, to mitigate this bias and achieve a fair assessment of performance,  we focus on the Matthews Correlation Coefficient (MCC) over different divisions of sequences that belong to the less frequent classes. We use two versions of MCC as measurements, namely MCC1, which contrasts specific signal peptides against TM/Globular (NO-SP) type proteins, and MCC2, which additionally includes all remaining sequences in the negative set.

\begin{figure}[H] 
\centering 
\includegraphics[width=1.0\textwidth]{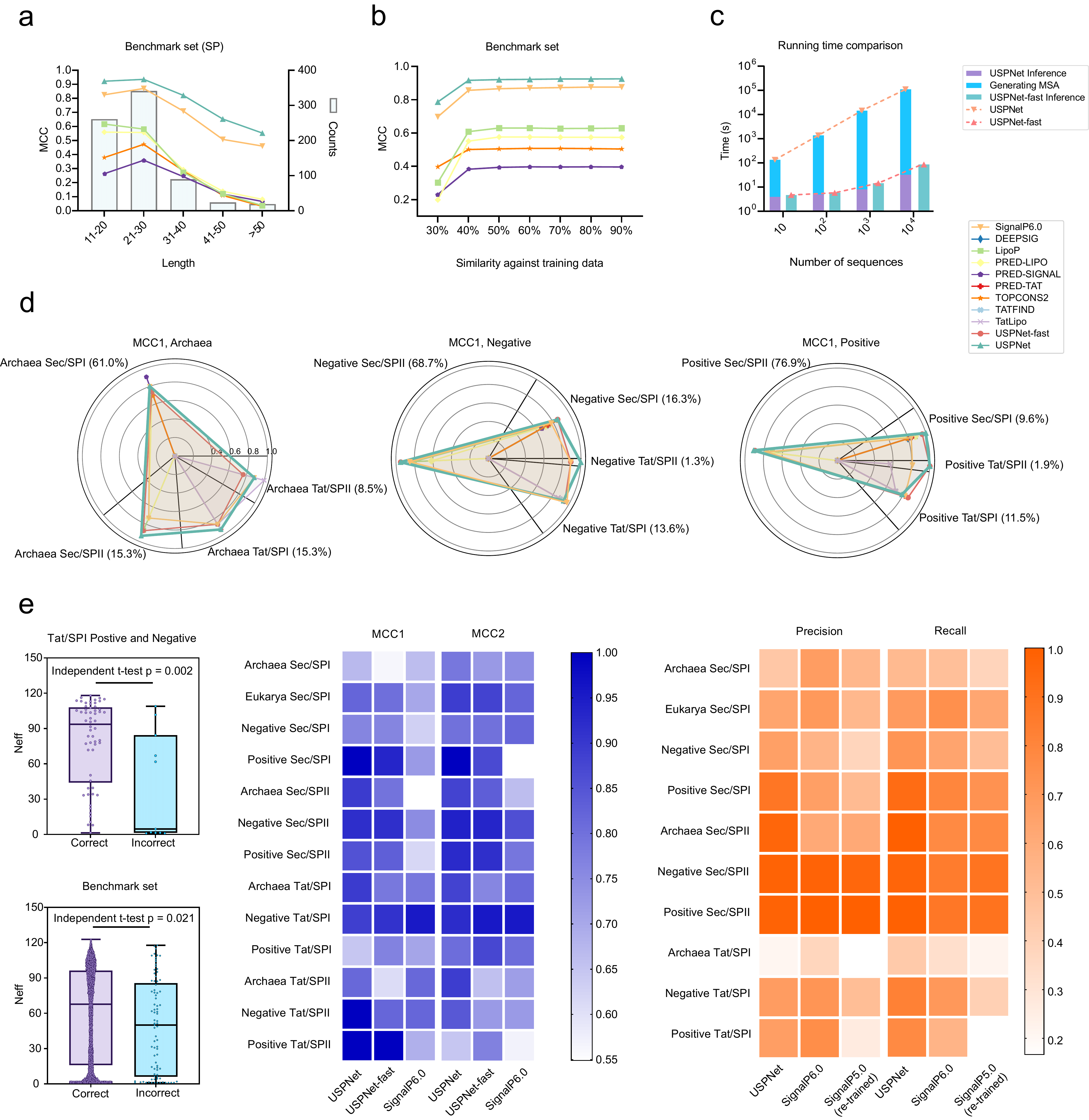}
\caption{\textbf{USPNet demonstrates robust performance across different signal peptide types and organism groups.} 
\textbf{a}. Length distributions with the performance of signal peptides in the benchmark set. Most SPs are below 50 AAs in length, and all models have degradation in performance when SPs are longer than 30 AAs.
\textbf{b}. The performance under different sequence identities between the benchmark set and training set.
\textbf{c}. Signal peptide prediction running time comparison between USPNet and USPNet-fast. The MSA generation step is time-consuming, accounting for roughly 95\% of the total processing time.
\textbf{d}. Radar charts of USPNet and other models on different perspectives of performance. We compare MCC1 of different models from the viewpoint of organism groups with proportional information of each SP type. USPNet behaves as the most powerful signal peptide predictor in every organism group. 
\textbf{e}. Comparison between USPNet, USPNet-fast, SignalP6.0, and SignalP5.0 on both signal peptide type prediction and signal peptide cleavage site prediction. And Neff scores of MSA on Tat/SPI of Gram+ and Gram-, and benchmark set. (Independent t-test: two independent samples t-test).} 
\label{dataset} 
\end{figure}


\begin{table}[h]
\centering
\caption{Statistics for the composition of the full dataset adopted in this study, with the benchmark dataset numbers in parentheses. SP denotes Sec/SPI, L denotes Sec/SPII, P denotes Sec/SPIII, T denotes Tat/SPI, TL denotes Tat/SPII, and N/C denotes TM/Globular(NO-SP) here.}
\label{DSTotal}
\scalebox{0.8}{
\begin{tabular}{lllllllll}
\hline
Dataset                                          & Organism      & SP   & L  &P  &T &TL     & N/C   & Total \\ \hline
\multicolumn{1}{l|}{\multirow{4}{*}{SP type prediction}} & Eukaryotes    & 2040 (146) & -   & -  & - & -  & 14356 (5581) & 16396 (5727) \\ \cline{2-9} 
\multicolumn{1}{l|}{}                            & Gram-positive & 142 (15)  & 516 (120)  & 4 (0)  &39 (18) &8 (3)   &226 (81) &935 (237)   \\ \cline{2-9} 
\multicolumn{1}{l|}{}                            & Gram-negative & 356 (61)  & 1087 (257) & 56 (0) & 313 (51)   & 19 (5) & 933 (133) & 2764 (507)  \\ \cline{2-9} 
\multicolumn{1}{l|}{}                            & Archaea       & 44 (36)   & 12 (9)  & 10 (0)   & 13 (9)   & 6 (5) & 110 (81) & 195 (140)   \\ \hline
\end{tabular}}\\[-2.4mm]
\end{table}

We compare signal peptide classification performance over different organism groups of several known signal peptide prediction models on the benchmark dataset. 
In total, 16 methods are selected for comparison; among them, SignalP6.0 and SignalP5.0-retrained are trained on the same training dataset as ours. 
Results for other methods are obtained directly from their publicly available web servers, which leads to potential performance overestimation due to the lack of homology partitioning. As the task at hand involves multi-class classification, we dissect the results based on the types of signal peptides. For the data with Sec/SP\uppercase\expandafter{\romannumeral1} labels, our model outperforms other methods on nearly all the organisms and metrics. The only exception is PRED-SIGNAL, specifically designed to detect only Sec/SP\uppercase\expandafter{\romannumeral1} SPs, which marginally outperforms us on MCC1 of Archaea. Nonetheless, our model still demonstrates superior performance on MCC2 (Figure \ref{dataset}.d and Supplementary Figure 3). When considering Sec/SP\uppercase\expandafter{\romannumeral2} SP classification, only 5 methods demonstrate capability, with corresponding data available for three organism groups. USPNet unequivocally has the best performance across all the metrics. For the MCC1 of Archaea, we exceed others by at least 6\%. Moreover, even our USPNet-fast does better than all other competitors. 

When looking at the performance from the perspectives of SP length and sequence identity, we find that detection performances for short SPs are good, and all methods have degradation when SPs are longer than 30 AAs (Figure \ref{dataset}.a). And when sequence identities are lower than 40\%, USPNet has minimal performance degradation (Figure \ref{dataset}.b), showing better generalization to proteins distant from the training data.
It is obvious that USPNet has made an impressive promotion in the prediction performances, especially over the minor classes. 

We also conduct a head-to-head comparison of MCC1 and MCC2 between USPNet and SignalP6.0, as shown in Figure \ref{dataset}.e.
USPNet achieves significant improvement on most SP types. Especially for data with Sec/SP\uppercase\expandafter{\romannumeral2} labels, we observe an increase of 10.0\% across all 4 groups. And on Tat/SP\uppercase\expandafter{\romannumeral2}, the class with the least data, USPNet is able to retrieve nearly all SPs. Despite SignalP6.0 demonstrating commendable performance, it falls short when compared with USPNet. On Tat/SP\uppercase\expandafter{\romannumeral1} of Gram-positive and Gram-negative, however, USPNet is behind USPNet-fast and SignalP6.0.  This can primarily be attributed to the subpar quality of multiple sequence alignments (MSA) in incorrect predictions. We calculate the number of effective sequences (Neff) of correct predictions and incorrect predictions used by the MSA-transformer in our method(Figure \ref{dataset}.e). Specifically, for Tat/SP\uppercase\expandafter{\romannumeral1}, there exists a significant Neff gap between correct and incorrect predictions: the median of the former is around 90, while the latter is only 4. Across the entire benchmark set, the MSA quality of correctly predicted SPs is also better, suggesting that high-quality MSA generation could boost the performance of USPNet. Overall, the results on the benchmark set accord with our assumption that the margin-based loss formulation allows the classification boundary of rare classes to be extended further and avoid overfitting in some ways. With better generalization, USPNet is an unbiased multi-class SP predictor that is able to predict all 5 kinds of SPs. Besides the benchmark set, we also conduct 5-fold cross-validation on the full training and benchmark sets (Supplementary Table 5-9). Since homology partitioning is inoperative in cross-validation, USPNet showcased more impressive performance, especially for the minor classes such as Tat/SPII and Sec/SPIII signal peptides.

\subsubsection*{Signal peptide cleavage site prediction}
Generally, the key focus of USPNet is to produce a predictor to classify various signal peptide types with a better generalization of different classes. However, before categorizing SPs, finding out the precise cleavage site of a protein is also an essential step in a real-world application. Here in USPNet, we utilize the attention weights of the context attention matrix to acquire information related to per amino acid, subsequently informing cleavage site decisions. For comparison, we employ both precision and recall rates to evaluate USPNet's performances on cleavage site prediction against SignalP6.0 and SignalP5.0-retrained, and the result can be observed from Figure \ref{dataset}.e. We notice that the overall performances on cleavage site predictions of USPNet are comparable to SignalP6.0. In especial, our model significantly outperforms the other 2 models on recall, which indicates that we are better at finding as many signal peptides in organisms as possible, especially those belonging to elusive minor classes. However, our model's precision is slightly lower than that of SignalP6.0 in certain classes, potentially due to the amplified weight allotted to minor classes, leading to a potential increase in false positives.  Nevertheless, users can adjust weights according to their specific needs, enhancing USPNet's effectiveness and suitability for their application scenarios.

\subsubsection*{USPNet versus USPNet-fast}

In the study, we introduce two versions of USPNet for use. They are USPNet and USPNet-fast. The rationale behind developing USPNet-fast is to provide a quicker prediction option, recognizing that the multiple sequence alignment (MSA) steps can be particularly time-consuming.  Inspired by this need, we incorporate ESM-1b to replace MSA-transformer and expedite prediction while retaining other modules (Figure \ref{main}.c). ESM-1b encodes properties of protein sequences across different scales. As confirmed in previous research, the model can learn structures of amino acids, protein sequences, and evolutionary homology solely from sequence data, without additional biological signals \cite{r16}.  To measure the degree of speed increase, we randomly select some sequences to test the prediction time of both USPNet and USPNet-fast (Figure \ref{dataset}.c). It shows the average inference time for the two methods is roughly the same. However, generating MSA is a demanding task, accounting for approximately 95$\%$ of the total processing time.  USPNet-fast, by avoiding the need for MSA files and working directly with amino acid sequences, is significantly faster. We evaluate the performance of USPNet-fast on the benchmark set and compare it with USPNet (Figure \ref{dataset}.e). On most categories of signal peptides, USPNet-fast is slightly behind USPNet, but the gap is modest. Notably, for Tat/SPI of both the Gram-positive and Gram-negative groups, USPNet-fast outperforms the standard USPNet by approximately 2\% and 7\%, respectively. Hence, when the MSA quality is not satisfying, applying USPNet-fast instead of USPNet is preferable. Generally, the fast version outperforms most previous methods (Figure \ref{dataset}.d) and offers substantial speed benefits.

\subsection{Protein language models and LDAM loss improve the generalization of USPNet}

In this part, we will discuss the performances of models with different loss functions and embeddings to look deeply into the reasons why USPNet is universal. We will compare LDAM loss with other commonly-used loss functions and analyze the performance differences between models with and without MSA and ESM embeddings. Notably, we employ overall MCC, Kappa, and balanced accuracy for evaluation instead of using MCC over different classes and organism groups. Because performance across minor classes for some models is remarkably similar, and these metrics can make overall performance more straightforward and intuitive to display differences.

\begin{figure}[H] 
\centering 
\includegraphics[width=1.0\textwidth]{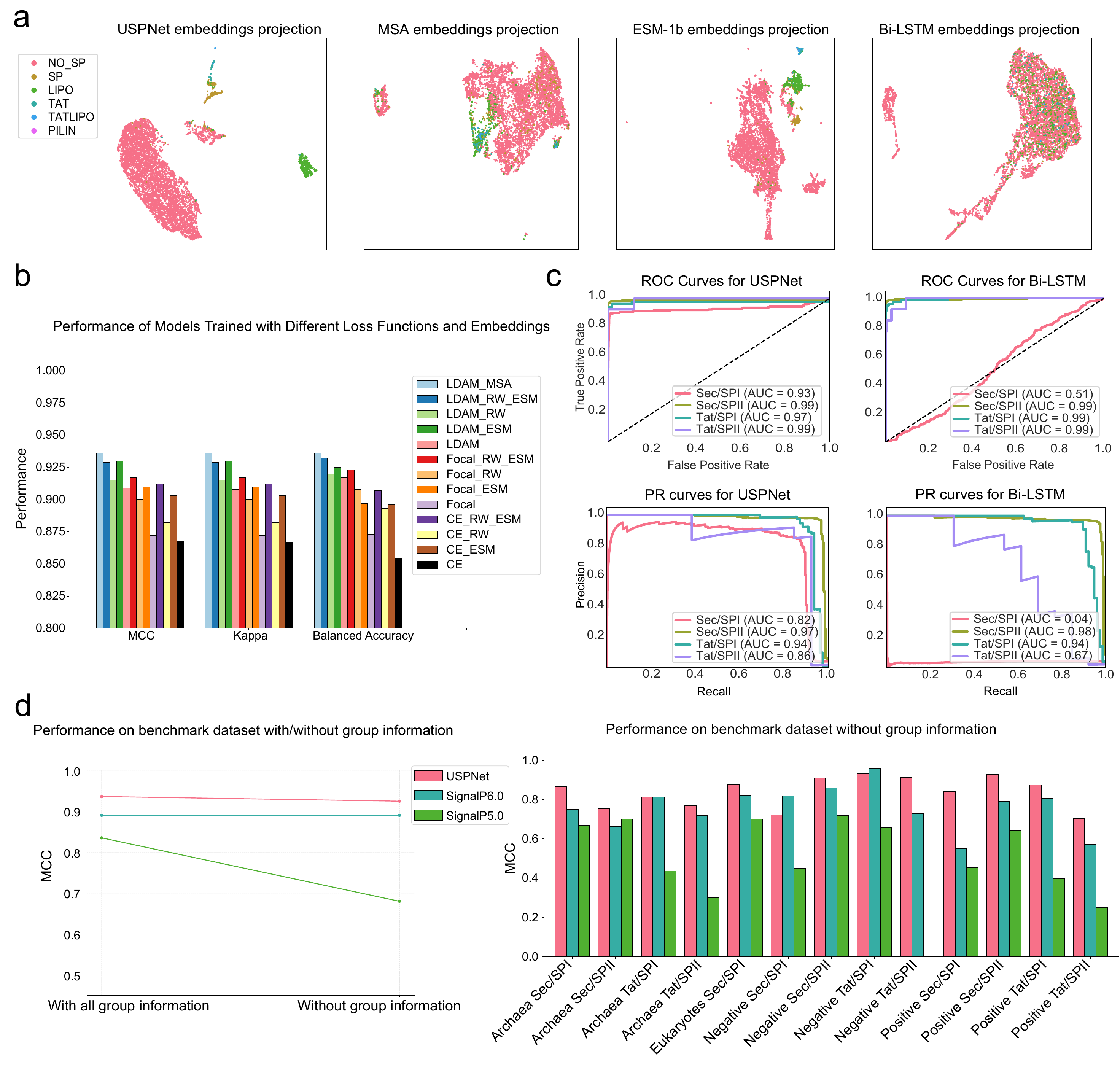}
\caption{\textbf{Embedding and ablation study performance analysis of USPNet compared to alternative models.} \textbf{a}. 2-D UMAP projection of different embeddings on benchmark data.  USPNet embeddings show the same types of peptides cluster together. Bi-LSTM embeddings are the last layer output of the Bi-LSTM model. Even if it is trained on SP data, we cannot see clear clusters of different SP types. Both protein language models (MSA-transformer and ESM-1b) form clusters between different SP types, indicating that they encode some functional information related to signal peptides. \textbf{b}.  Ablation study performance of USPNet: MCC, Kappa, and Balanced Accuracy of models trained with different loss functions and embeddings. \textbf{c}. ROC curves and  Precision-Recall curves of USPNet and Bi-LSTM on different types of signal peptides prediction. Our model is stable across all 4 kinds of SPs, while Bi-LSTM suffers from the data imbalance problem.  \textbf{d}. Comparison of USPNet's organism-agnostic performance against SignalP6.0 and SignalP5.0 on the benchmark dataset, when organism group information is missing, showcasing comparable performance to SignalP6.0 without obvious performance degradation and significant outperformance compared to SignalP5.0 across all categories. } 
\label{embeddings} 
\end{figure}

Before delving into the experiment results, to verify the efficiency of protein language models sequence embeddings, we perform visual processing of the embedding vectors on our benchmark data. We extract embeddings from 4 models and apply UMAP \cite{mcinnes2018umap} for data dimensionality reduction. UMAP uncovers similarity patterns in data points with multiple features, allowing us to observe the convergence of data points with high similarity into clusters when the dimensionality-reduced results are projected onto a two-dimensional plane. For USPNet, we extract embeddings
from the last embedding layer, the result is shown in Figure \ref{embeddings}.a. We can observe some obviously differentiated clusters, and each cluster mainly consists of one type of data, which indicates USPNet nicely mines the functional features inside the peptide sequences and is efficient in classifying different types of signal peptides. For the two protein language models applied in our study to build USPNet and USPNet-fast, we take their output directly for visualization. As these two language models are trained on large-scale protein data via self-supervised learning, they have learned the structural and functional information of universal protein sequences. And results show that the models also encode the functional information of signal peptides. At last, to evaluate the help of protein language models from another perspective, we train a model that removes the embeddings of protein language models and remains all other modules, which we call Bi-LSTM. Although Bi-LSTM is trained on the same data as USPNet, it cannot splendidly discover the pattern of signal peptides and therefore does not encode much valuable information into its embeddings. We then evaluate the efficiency of different modules in USPNet by concrete measurable metrics.

We first trained the USPNet model with three loss functions: cross-entropy function, focal loss, and LDAM loss \cite{r23}. The other two introduced here are for better comparability to display the advantages of LDAM loss. 
The hyperparameters of loss functions are chosen manually according to multiple runs. In Figure \ref{embeddings}.b, by analyzing the results of experiments, it is clear that LDAM loss allows for maximum gains in improving the accuracy of predictions. With the same model architecture and ESM embedding, the LDAM\_ESM model trained based on LDAM loss reaches the MCC of 0.93, which is 1.8\% higher than the Focal loss and 2.6\% higher than cross-entropy loss, respectively. In addition, we try to apply the deferred re-balancing training procedure proposed with the LDAM Loss (LDAM-RW). However, when training the signal peptide predictor, the LDAM-RW algorithm cannot help gain benefits. Instead, we propose to directly apply the combination of LDAM Loss and class-balanced loss\cite{r25}, called CB-LDAM loss. Except for baseline loss functions, we used focal loss and cross-entropy loss with reweighting for comparison. We manually give different weights according to the frequency of different signal peptide labels to enhance generalization. And these manual reweighting factors are acquired based on the results of multiple runs. Results show that class-balanced reweighting from the start will help models make further progress.

Finally, to evaluate the functionality of protein language model embeddings, we add one, two, and three linear layers, respectively, to integrate information from original ESM or MSA embeddings. The results of experiments suggest that we get the best results using two layers. And the reason for the decline in performance with more layers may be overfitting. It is clear that the model can acquire continuous improvement by integrating information from state-of-the-art embeddings. Compared with the ESM embedding, the MSA embedding further improves the overall performance of the model. Besides, we also remove the MSA or ESM embeddings to evaluate the performance of Bi-LSTM. Specifically, we take AUCROC and AUCPR
to make the comparison (Figure \ref{embeddings}.c). USPNet behaves steadily on all types of signal peptides and keeps the value of AUROC above 0.9 and AUPRC above 0.8. Interestingly, the lowest performance is seen in class Sec/SP\uppercase\expandafter{\romannumeral1}, which indicates USPNet's excellent performance for minor classes. But when we look into the Bi-LSTM, it shows that the performance is extremely poor on type Sec/SP\uppercase\expandafter{\romannumeral1}. One reason is that we set the weight of the minor classes at a high level, making the model inclined to predict the signal peptides as minor classes. However, if we do not set the weights, the model can suffer from data imbalance problem \cite{r17}, compromising its ability to identify these minor classes. In summary, the protein language models boost the generalization, while our loss function strategically shifts the model's focus towards minor classes. Working in unison, these factors enable USPNet to classify signal peptides precisely and resolve the data imbalance.

\subsection{USPNet is robust against the organism-agnostic experiments}

The previous results indicate USPNet is able to make predictions that outperform other previous models in classifying signal peptide types, especially in the minor classes.
To further tap the group-information-independent capacity of USPNet, in this part, we carry out organism-agnostic classification.

Most previous proposed signal peptide prediction tools, including SignalP5.0 \cite{r14} and TargetP2.0 \cite{r17} rely on extra group information of sequences to enrich embeddings and boost their performance. Typically, this involves incorporating a one-hot embedding vector with a specific number of dimensions corresponding to group origins into the model.
However, in the context of metagenomic research and various other applications, obtaining group information for all sequences is not practical. It is thus essential to develop methods capable of accurately predicting signal peptide types using solely amino acid sequences.
In the following experiment, we still utilize the benchmark dataset, but remove the organism group information. Specifically,  group information is missing in the input for USPNet.  We also include the retrained SignalP5.0 for comparison since SignalP6.0 does not rely on group information \cite{signalp6.0}. For SignalP5.0, we input randomized group information instead because it does not support blank input of group information. The performance is assessed using the Matthews Correlation Coefficient (MCC2).

As depicted in Figure \ref{embeddings}.d, remarkably, USPNet's performance, in general, remains high and exhibits no significant degradation in signal peptide detection without group information. A comparison of the overall performance (multi-class MCC) between the three models reveals that USPNet still behaves the best when missing group information. This demonstrates the resilience and adaptability of USPNet in scenarios where group information is unavailable.

Conversely, SignalP5.0 experiences a substantial decline across all categories when group information is absent. Specifically, SignalP5.0 displays a notable decrease of 0.155 in its multi-class MCC when without group information. The signal peptides of minor classes concentrate on Archaea, Gram-Negative, and Gram-Positive groups, which suffer the most when group information is missing.

One reason for the robust performance of USPNet is that protein language models \cite{r16, MSA} applied in USPNet are able to encode protein sequences into a high-level feature space and represent biological variation. The two models take advantage of enormous protein data and learn structural/functional properties as well as evolutionary information. The representations will possibly substitute for the formerly mentioned group embedding. With our manipulation that concatenates MSA/Evolutionary Scale Modeling (ESM) embeddings with the Bi-LSTM embeddings, USPNet is turned into an organism-agnostic model to handle the universal database reliably. What's more, USPNet has the potential to serve as an end-to-end model to directly get results based on residue-level sequences from unknown origins, which can be beneficial for typical cases of metagenomic data.

\subsection{USPNet has impressive generalization ability and works well on domain-shifted data}
\subsubsection*{Performance of USPNet on the rigorous non-retrospective independent test set }

USPNet performs well on the benchmark set and does not rely on group information. However, sound performance may be expected on account that it is done in the same domain as the training set. In order to verify our model's robustness in a domain-shift dataset, we first curate an independent test set, called SP22, with stringent criteria to ensure its differentiation from our training dataset (40\% sequence similarity, published after November 2020, i.e., the date of benchmark dataset collection, and so on, see Methods). SP22 comprises 43 protein sequences from 39 species, which are largely absent from the training set, with each sequence containing a signal peptide. Among them, 31 are from Eukaryotes group, 6 are from Gram-negative group, and 6 are from Gram-positive group.

\begin{figure}[H] 
\centering 
\includegraphics[width=1.0\textwidth]{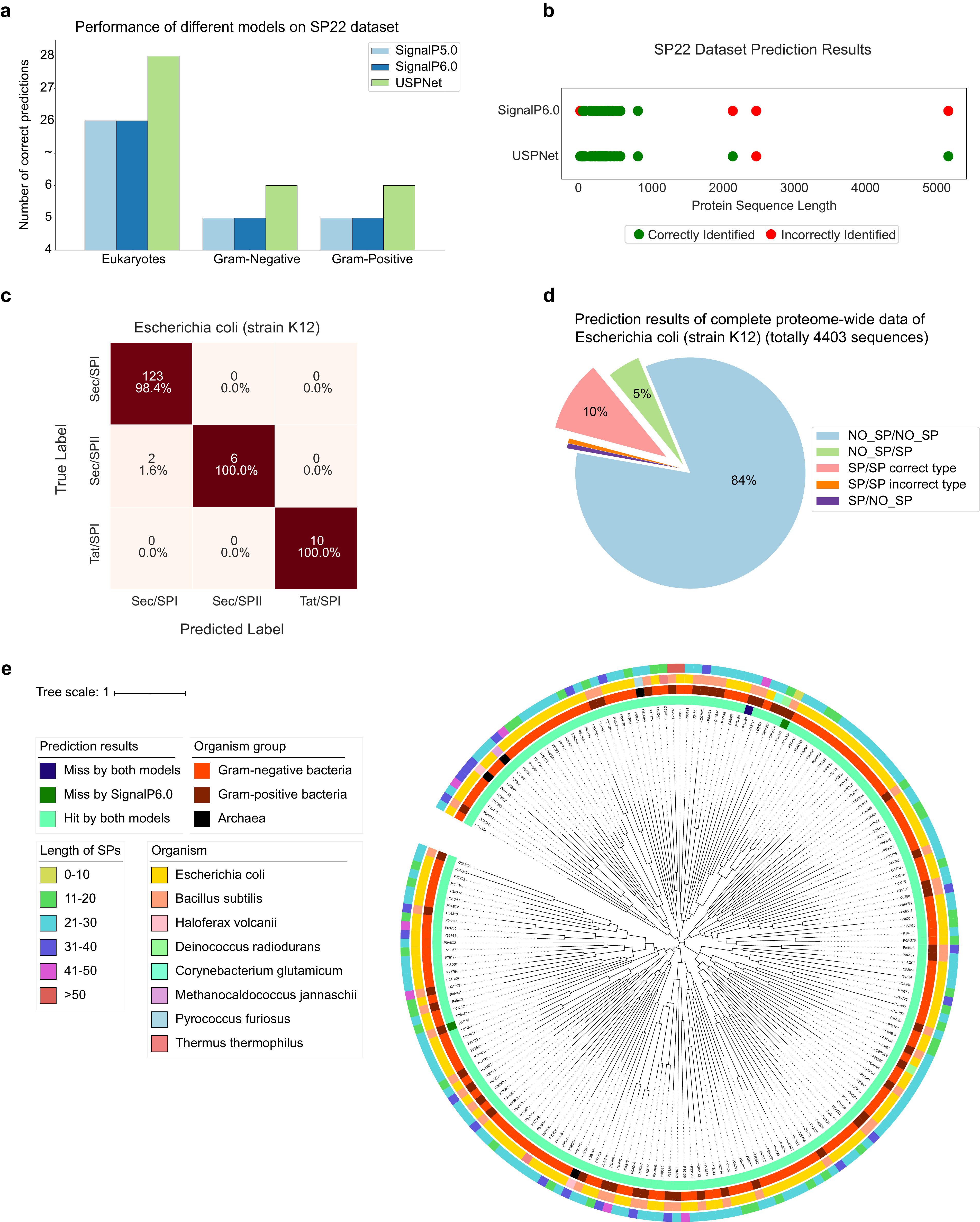}
\caption{\textbf{Performance of USPNet on domain-shift data.} \textbf{a}. The rigorous non-retrospective SP22 independent dataset performances of different models on sequences from Eukaryotes, Gram-Negative, and Gram-Positive groups. \textbf{b}. Scatter plot of the prediction of USPNet and SignalP6.0 on the SP22 dataset. USPNet can retrieve the signal peptides in both very short and long proteins. 
\textbf{c}. Escherichia coli (strain K12) proteome-wide prediction performance of USPNet. It accurately classifies almost all proteins.  \textbf{d}. The UniProt labels and USPNet prediction results' matching conditions on the complete proteome data of Escherichia coli (strain K12).
\textbf{e}. Cladogram of species tree of all the 193
reference proteomes with experimentally verified SPs. The annotation rings from inner to outer are: 1) prediction results of USPNet and SignalP6.0; 2) organism group of proteins; 3) detailed organism of proteins; 4) length of signal peptides.} 
\label{SP22} 
\end{figure}

We analyze the performances of USPNet as well as SignalP6.0 and SignalP5.0 on the SP22 dataset to see if our model works well on domain-shift data. We collect the results of SignalP6.0 and SignalP5.0 by feeding sequences into their web servers. USPNet correctly identifies 28 Eukaryotes SP sequences, which is better than SignalP6.0 and SignalP5.0, as Figure \ref{SP22}.a shows. For groups of Gram-negative and Gram-positive, USPNet correctly identifies all 6 Gram-negative SP sequences and 6 Gram-positive SP sequences, while both SignalP6.0 and SignalP5.0 mispredict one SP sequence for each organism group. We then analyze the misclassified protein sequence made by the two models (Figure \ref{SP22}.b) and find that USPNet can correctly predict the SP type of two particularly long protein sequences, one with a length of 5206 in the Gram-negative group and one with a length of 2178 in the Gram-positive group, while other models can not. Additionally, USPNet can classify the shortest sequence in the SP22 dataset, with a length of 36 from the Eukaryotes group. The results demonstrate that USPNet is able to handle proteins with various lengths, while previous methods have exhibited limitations in this aspect. The performance on the SP22 further unveils the advantages of USPNet in learning features of data from various sources with minor-class labels and the ability to make unbiased predictions.

\subsubsection*{USPNet retrieves most SPs on proteome-wide data}

Besides the independent test, we want to evaluate if USPNet is effective with the case study for species.  Consequently, we examine the proteome-wide prediction performance of USPNet in Escherichia coli (strain K12) as well as other 7 organisms and compare it with the benchmark model, SignalP6.0. The data we selected comprise the entire set of proteins that can be expressed by an organism. To ensure accuracy in subsequent classification, we demand high-resolution cleavage site prediction. Despite the variability in protein lengths, our restricted input makes USPNet focus on analyzing the N-terminus of a sequence and therefore alleviates the influence of long proteins. Moreover, the attention mechanism enables the holistic analysis of the input sequences and thus makes USPNet performs meticulous detection.

We further assess the proteome-wide performance using the well-annotated proteome of Escherichia coli (strain K12) (Figure \ref{SP22}.c). 
Out of the 141 experimentally verified SPs, 125 are Sec/SPI, 6 are Sec/SPII, and 10 are Tat/SPI SPs. USPNet successfully detects all 141 experimentally verified SPs. Out of them, USPNet accurately predicts 123 out of 125 SPs as Sec/SPI type and precisely predicts all 16 SPs from Sec/SPII and Tat/SPI type. As one of the most advanced methods, SignalP6.0 also predicts all SPs from strain K12.

Besides the experimentally verified signal peptides, we also apply USPNet to the complete proteome-wide data of Escherichia coli (strain K12). Specifically, we use the reference proteome with Proteome ID
UP000000625 \cite{blattner1997complete}, which includes the complete genome sequence of Escherichia coli K-12, as the dataset for our study. The dataset contains 4403 protein sequences. Among them, 3907 are annotated as NO-SP and 496 are annotated as SP by UniProt. USPNet predicts 3734 proteins as NO-SP and 669 as SP. The detailed matching information is shown in Figure \ref{SP22}.d, before `/' is UniProt's annotation, and after `/' is USPNet's prediction. `SP/SP correct type' means the type of predicted SP is the same as the annotation, while `SP/SP incorrect type' means the predicted SP type and annotation are different. Notably, most of the prediction results are matched with the annotations, and only around 6\% are inconsistent. However, it is difficult to verify the correctness of the result because some annotations in UniProt are automatic annotations provided by existing prediction tools. In other words, some of them are not experimentally verified. Nonetheless, USPNet's genome-scale predictions yield estimates of plausible results.

For other reference proteomes from 7 different organisms, the UniProt database reports 52 experimentally verified SPs. USPNet shows a strong performance that seeks out 51 experimentally verified SPs, with only one miss prediction in Bacillus subtilis. However, in comparison, SignalP6.0 makes 3 false predictions for experimentally verified SPs from Bacillus subtilis. We then reconstructed the phylogeny of all proteins and obtained the unrooted tree, which is annotated with the prediction results (Figure \ref{SP22}.e). The incorrect predictions made by both models are all proteins from Gram-positive bacteria.
The result shows the effectiveness of USPNet in finding out proteome-wide signal peptides across different organisms with greater accuracy than previous methods. The near-perfect performance demonstrates USPNet can potentially handle the annotation of proteomes with high confidence when there is no experimental validation.

\begin{figure}[H] 
\centering 
\includegraphics[width=0.89\textwidth]{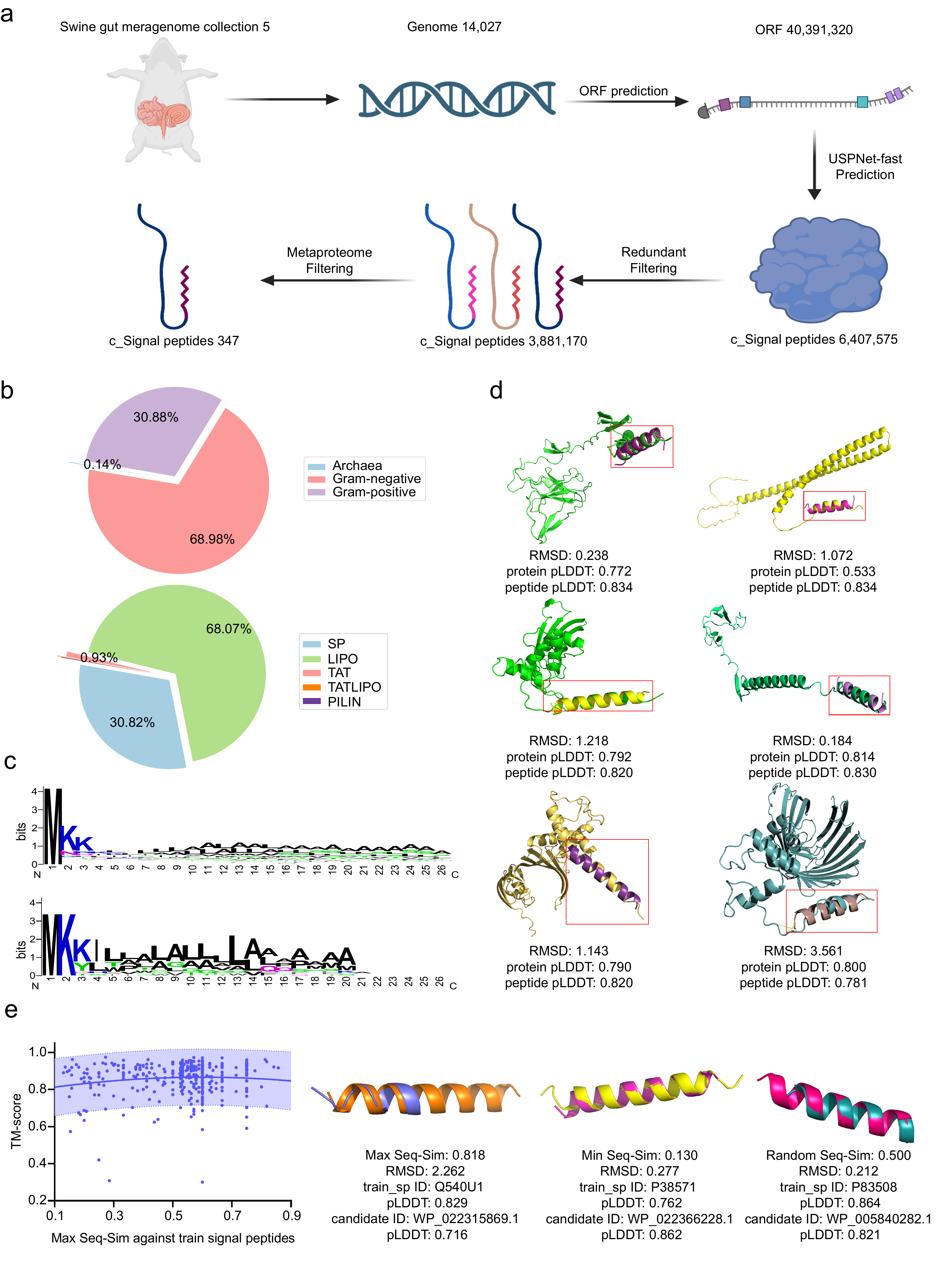}
\caption{\textbf{The exploration of metagenomics data for signal peptide discovery. }\textbf{a}. Schematic representation of the USPNet pipeline for novel signal peptide discovery in metagenomics data. 347 novel SP candidates were identified. \textbf{b}. Pie charts depicting the distribution of 3,881,170 protein sequences predicted by USPNet that contain signal peptides, categorized by predicted SP types (left) and organism groups (right). \textbf{c}. Sequence logo for 347 novel signal peptide candidates (top), as well as 16 artificial novel SPs (bottom). Two sets of peptides exhibit a similar residue distribution. \textbf{d}. Visualization of the alignment between select proteins and their predicted signal peptides, accompanied by RMSD and pLDDT values; the majority exhibit alignment with the predicted novel SPs. \textbf{e}. Distribution of the highest sequence and structure similarity between 347 candidates and 4 experimentally verified SPs mentioned in our study to that in the training dataset. The sequence identity hit a low of only 13\%, but the structural similarity is strikingly high, with TM-scores mostly above 0.8. And 3 pairs of sequences, one with maximum sequence similarity, one with minimum sequence similarity, and a random one, are sampled and aligned. They have analogous structures.} 
\label{agnostic} 
\end{figure}

\subsection{USPNet enables the discovery of novel SPs based on structural information with extremely low sequence similarity}

The benchmarking exhibits that USPNet enables high-resolution signal peptide prediction, and drastically improves performance on the minor classes compared with previous methods. Besides, the proteome-wide research demonstrates our method effectively detects the SPs from the proteomes of various organisms. Moreover, in the organism-agnostic experiment, the performance of USPNet seems to be group-information-independent. All the above studies provide evidence
that USPNet has the potential to discover SPs from multiple sources of data. One of the most preliminary data for research is metagenomics. However, previous works have not shown that they can efficiently detect SPs or even novel SPs from original metagenomics data. At the same time, no standard pipeline is available for signal peptide discovery from metagenomics cohorts. Therefore, we would like to evaluate if UPSNet is applicable in real-world metagenomics research and explore the pattern of USPNet in recognizing SPs. In this section, we build a complete pipeline from collecting metagenomic data to make novel signal peptide detection. The pipeline is shown in Figure \ref{agnostic}.a. We studied the swine microbiome, which has a complex community structure. And its constituent microbes, especially those in the gut, could employ many functional proteins like AMPs \cite{ma2022identification} to help compete for resources or stabilize the community structure. Many of these proteins are likely dependent on signal peptides for their functionality, providing an adequate resource for finding SPs. 

Here we collect the swine gut metagenomic data from five projects and resources. In total, we have 14027 genomes, which are used to predict ORF sequences. To generate high-quality protein sequences, we first integrate fastp \cite{chen2018fastp} as a quality controlling tool and utilize PLASS \cite{steinegger2019protein} to assemble nucleotide sequences to protein sequences, which leads to 40,391,320 sequences. Then, considering the amount of data as well as the time cost, we apply USPNet-fast to recognize signal peptides on all sequences. As a result, 6,407,575 are predicted as signal peptides (c\_Signal peptides). The total running time of USPNet-fast for completing all predictions is 211,932 seconds. To avoid some potential false positives, we remove the genomes that are without organism group information, and the known signal peptides are also removed. Then, we get 3,881,170 sequences. The distribution of 3,881,170 protein sequences according to predicted SP types and group information is shown in Figure \ref{agnostic}.b. The majority of sequences come from gram-negative bacteria. Sec/SPII type SPs are predicted to be the most frequent in both Gram-negative and Gram-positive bacteria and second frequent in Archaea. Sec/SPI type SPs are predicted to be the most frequent in Archaea. 

Given the known effect of signal peptides in directing the transfer of synthesized proteins to the secretory pathway, we further select sequences that are likely expressed to proteins/peptides. We accomplish this by identifying the same peptides in swine gut metaproteomics data, and eventually identify 347 sequences. Besides novel peptides, our method also retrieves all the 4 experimentally verified signal peptides that exist in swine gut data but do not exist in the previously mentioned datasets (Table \ref{10_1732}). The evidence shows their exhaustive functions in controlling protein secretion and translocation, which belong to different families and play a part in different organisms. Therefore, our comprehensive filtering guarantees all 347 peptides are novel in comparison with experimentally verified SPs. We compare them with 16 artificially designed and synthesized novel SPs, which have various compositions in the h-region and are validated to improve the secretion \cite{han2017novel}, by logo plot (Figure \ref{agnostic}.c). Notably, our candidate peptides show similar motifs to these novel signal peptides: the second and third places are dominated by lysine. For the rest of the positions, alanine and leucine take first place. The similarity between these sequences indicates they may have homologies. We also applied USPNet on these 16 novel SPs. It precisely classifies all their types and cleavage sites. To look for further evidence that our discovered peptides are probably SPs, we search in the UniProtKB \cite{uniprot2023}. It shows that 1,715 proteins contain these peptides with a confident label ECO:0000256 (automatic annotation), demonstrating that they are automatically classified as signal peptides by the database with high confidence, 10 most frequently occurring sequences of 347 predicted novel signal peptides are provided in Table \ref{10_1732}. The most often seen sequence is WP\_008666378.1 (NCBI ID), which appeared 179 times, about one-tenth of all proteins. In total, we have 281 peptides that appeared in the 1715 sequences. We then choose some of these proteins and predict their 3-D structure using Alphafold2 \cite{jumper2021highly}. Thereafter, we align them with their corresponding predicted signal peptides (Figure \ref{agnostic}.d). Prominently, the N-terminus sequences of these proteins are mostly well-aligned with our predicted novel SPs. Meanwhile, the structural connection between such sequences and their subsequent sequences does not seem very strong. It signifies that these N-terminus sequences are more likely to be deleted or replaced when functioning, which coincides with how signal peptides work. Finally, our analysis reveals that the maximum sequence identity between our candidate peptides and the closest signal peptides (SPs) in the training dataset is merely 81.8\%. The majority exhibits less than 60\% similarity, with the lowest similarity being only 13\%. However, their structures have a remarkable likeness, the average TM-score is above 0.8. We further perform structural alignment on 3 pairs of sampled sequences with the highest similarity value, the lowest similarity value, and the random one (Figure \ref{agnostic}.e). These results further underscore the structural homology between our candidate peptides and specific signal peptides. Our method takes raw amino acid sequences as input and does not rely on structural information. However, the discovered candidates seem to have been screened out based on structural similarities. To further test our hypothesis, next, we look into the 4 verified SPs. Our original training set contains SPs exhibiting over 60\% similarity to these verified SPs. To ensure USPNet detects SPs without depending on sequence similarity, we exclude SPs from the training set with more than 40\% sequence identity to 4 verified SPs and retrain our model. This time, our model is still able to detect all 4 SPs, which on average, have quite high structural similarities to the training set (Table \ref{10_1732}).
The literature-based evaluation and database search results show that our USPNet is scalable to large amounts of data. Although it is a sequence-based method, it can discover peptides that have low sequence identity but high structure resemblance, which shows that our method has learned to represent protein folding information without any related inputs at a relatively ideal speed. It may potentially help to screen signal peptides from metagenomics data and discover SPs that differ significantly from what is currently known.

\begin{table}[!t]
\centering
\caption{Top 10 most frequently occurring predicted signal peptide sequences in 1715 proteins (Novel SP candidates), and some experimentally verified SPs that we figure out from swine gut metagenomics data. (Seq-sim: highest sequence similarity value in the reconstructed training set, TM-score: highest structure similarity (TM-score) value in the reconstructed training set)}
\label{10_1732}
\scalebox{0.9}{
\begin{tabular}{llrll}
\hline
\multicolumn{5}{l}{Novel SP candidates}                                                      \\ \hline
protein ID (NCBI)               & Predicted SP                 & Count & \multicolumn{2}{l}{Evidence}            \\ \hline
WP\_008666378.1                 & MNKKFLSAILFGALMVSSTGT        & 179   & \multicolumn{2}{l}{\multirow{10}{*}{\begin{tabular}[c]{@{}l@{}}ECO:0000256 \\ Automatic assertion \\ according to \\ sequence analysis. \\ SAM:SignalP\end{tabular}}} \\
WP\_008661887.1                 & MNKKFLSVILFSALMVGTAGT        & 102   & \multicolumn{2}{l}{}                        \\
ABG29383.1                      & MKKTLVSALTTALVVGAASTTFA      & 61    &  \multicolumn{2}{l}{}                   \\
WP\_009145071.1                 & MKKSLVLAMAMALGVTASA          & 49    &   \multicolumn{2}{l}{}                    \\
WP\_005825009.1                 & MNLNFRRTALLVGICSAVSLTYTPQLFA & 32    &   \multicolumn{2}{l}{}                  \\
WP\_006847594.1                 & MRRLTLLLVSLVLLSIQSVLA        & 32    &   \multicolumn{2}{l}{}                    \\
WP\_006846852.1                 & MKHLKQFLLMAMLLFTLAPLQVSA     & 30    &  \multicolumn{2}{l}{}                       \\
WP\_006848777.1                 & MKKIAILSLSLTLAAAAQA          & 28    &    \multicolumn{2}{l}{}                        \\
WP\_006846306.1                 & MYQQLKRASMALTLSSVCFLAFA      & 27    &  \multicolumn{2}{l}{}                      \\
WP\_005844474.1                 & MIKKLYLPLVALLVLALSS          & 26    &   \multicolumn{2}{l}{}                          \\ \hline
\multicolumn{5}{l}{Experimentally Verified}     \\ \hline
Entry (UniProtKB/Swiss-Prot ID) & Predicted SP                 & \multicolumn{1}{l}{Evidence}  & Seq-sim   & TM-score\\ \hline
P02768                          & MKWVTFISLLFLFSSAYS           & \multicolumn{1}{l}{\cite{patterson1977bovine}}  & 0.222 & 0.990\\
P28800                          & MALLWGLLALILSCLSSLCSAQ       & \multicolumn{1}{l}{\cite{christensen1992bovine}}  & 0.35 & 0.902  \\
Q2UVX4                          & MKPTSGPSLLLLLLASLPMALG    & \multicolumn{1}{l}{\cite{baldo1993adipsin}}  & 0.381 & 0.709   \\
Q3MHN5                          & MKRILVFLLAVAFVHA  & \multicolumn{1}{l}{\cite{nykjaer1999endocytic}}  & 0.125 & 0.961                                                              \\ \hline
\end{tabular}}
\end{table}

\section{Discussion}
We develop USPNet as a deep-learning method focusing on resolving the existing long-tail data distribution problem and group-information dependency in signal peptide prediction.
To solve the data imbalance, we first include LDAM, which is orthogonal to most deep-learning-based signal peptide predictors. As the single loss function has limited effects Based and considering the feature of LDAM, we propose to combine class-balanced loss and LDAM loss together in our method, which we named CB-LDAM loss to enhance the generalization of boundaries of minor classes and major classes. Furthermore, the developed protein language models incorporate rich evolutionary and structural information and have shown their excellent abilities to improve plenty of downstream tasks. Therefore, we propose to introduce MSA transformer and ESM embeddings to achieve an end-to-end organism-agnostic model. Finally, we present a new model architecture. The attention-based BiLSTM model allows for maximizing the possibilities to integrate and extract relationships among different positions of inputting sequences and helps boost performance. 

Although comprehensive experiments have verified that the integration of all these techniques can achieve better performance in signal peptide prediction compared with state-of-the-art methods, our result on recognizing Sec/SPI signal peptide of Archaea is not that impressive. We suppose the reason for this phenomenon is our loss function adjusts the weight of different types of SPs, and hence makes USPNet tend to classify peptides into minor classes. During real-life usage, to make the model more focused, we recommend that users can tune the weights of different types according to their demand.

As previously highlighted, we develop a complete pipeline to discover novel signal peptides from metagenomics data. The reason we choose USPNet-fast for the study is the multi-sequence alignment can be time-consuming, making it impossible to finish the large-scale screening in a reasonable time. The 347 candidate novel SPs are available as an open science resource. Potentially, new signal peptides may be used to improve the secretion efficiency of heterologous proteins. We provide a paradigm for the discovery process, and it is easy to reproduce. Besides swine gut data we used in our study, other resources like the human gut and cardiac tissue are also signal peptide reservoirs. We expect our pipeline can give new insights into the related research.

Apart from signal peptide prediction, data imbalance, and object—dependence is very common in biological prediction problems. Our model-developing strategy can be transferred to more general fields of bioinformatics \cite{li2018deepre,yu2021hmd,lam2019deep,wei2021protein, yu2021hmd, li2021hmd}.
In summary, USPNet represents a widely applicable framework for predicting signal peptides or even protein sequences. Considering the fact that it can be integrated with other tools in a pipeline, we believe USPNet will be helpful to investigate a wide range of signal peptide problems.

\section{Methods}
\subsection{Data collection}
We collected, in total, four kinds of datasets in our study, including benchmark data, independent dataset, proteome data, and metagenomic data.
\subsubsection*{Benchmark data}
The training and benchmark test set is the same as introduced in SignalP6.0 \cite{signalp6.0}, which reclassified some SP types for data published with SignalP5.0 \cite{r14}. Further, it added some new SPs from UniProt20 \cite{r19_} and Prosite21 \cite{sigrist2012new}, and new soluble and transmembrane proteins from UniProt and TOPDB22 \cite{dobson2015expediting}. Then, part of the data in the new dataset was removed with the homology partitioning methodology introduced by Gíslason et al. \cite{gislason2021prediction}. In total, the training data contains 13679 sequences, and the benchmark test data contains 6611 sequences.
The original training dataset comprises proteins from four organism groups: Eukaryotes, Gram-positive, Gram-negative bacteria, and Archaea. To verify the effects of better generalization on minor classes, we carry out five separate SP types: Sec/SPI, Sec/SPII, Tat/SPI, Tat/SPII, and Sec/SPIII SPs. Other proteins are accordingly considered as TM/Globular (NO-SP) type, as shown in Table \ref{DSTotal}. The dataset with 6 separate labels has a long-tailed label distribution, which means that the NO-SP type is superior in numbers among different labels. Notably, there are three minority class SPs: Tat/SP\uppercase\expandafter{\romannumeral1} contains 365 data points, and even worse, only 33 data are labeled as Tat/SP\uppercase\expandafter{\romannumeral2} signal peptides, and 70 are labeled as Sec/SP\uppercase\expandafter{\romannumeral3} signal peptides.
In both training and benchmark sets, the length of the protein sequences can be varied. However, the lengths of signal peptides usually fall in 10-50 AAs (Figure \ref{dataset}). We thus set a cut-off at 70 AAs to highlight signal peptides and avoid the influence of the long protein sequences.

\subsubsection*{Independent test set}

The independent test dataset we curated is named SP22. It collected verified SP proteins from the Swiss-Prot database \cite{r19_}, which was released after November 2020, i.e., the date of SignalP 6.0 dataset collection. Specifically, it is generated in the following steps: (1) we remove proteins before November 2020, and proteins composed of less than 30 amino acids; (2) we select those from
eukaryotes, Gram-positive and Gram-negative bacteria; (3) we
select proteins containing signal peptides with a confident label
ECO:0000269 (experimental annotation) and ECO:0000305
(manually curated annotation) from the Swiss-Prot database
(released on 2022 04). Furthermore, to better ensure the
independence of the SP22 dataset, CD-HIT \cite{cd-hit} is applied to
remove redundant proteins sharing more than $40\%$ similarity
with proteins in SiganlP6.0 dataset, and internal redundancy
of SP22 is cut off to $80\%$. In total, SP22 has 43 protein sequences
from 39 species that seldom appeared in the training set, and
all of the sequences have signal peptides. Among them, 31 are
from Eukaryotes group, 6 are from Gram-negative group, and
6 are from Gram-positive group.

\subsubsection*{Proteome-wide data}
We collect reference proteome data from UniProt database. For Escherichia coli (strain K12), UniProt reports 496 proteins with signal peptides, 141 of which are experimentally verified and 355 predicted in various ways. Experimentally verified SPs are collected as signal peptides with a confident label ECO:0000269 and ECO:0000305 from the Swiss-Prot database (released on 2022\_04) for all proteomes. We also study other 7 reference proteomes available in UniProt, including Bacillus subtilis, Corynebacterium glutamicum, Deinococcus radiodurans, Haloferax volcanii, Methanocaldococcus jannaschii, Pyrococcus furiosus, and Thermus thermophilus. 

\subsubsection*{Metagenomics data}
We collect metagenomes of swine gut from five resources: PRJEB38078 \cite{youngblut2020large}, PRJNA561470 \cite{looft2015complete}, PRJNA647157 \cite{zhou2020characterization}, CNP0000824 \cite{chen2021prevotella}, and Global Microbiome Conservancy (GMBC) \cite{groussin2021elevated}. They are used for predicting sORF sequences. And to ensure that our sORFs are indeed expressed, we used a metaproteome dataset from PRIDE project PXD006224 \cite{tilocca2017dietary}. We select sequences that are identical to our predicted signal peptides. 

\subsection{Establishment of USPNet}
Considering the length of signal peptides in protein N-terminus, we set a cut-off as 70 to the input sequence length of protein L, which means that each input sequence contains at most 70 amino acids. Considering the number of different residue types is 20, we employ an embedding layer to extract features and therefore convert input sequences into the L×20-dimensional matrix. At the same time, we use one-hot encoding to express 4 kinds of group information (Eukaryotes, Gram-positive, Gram-negative bacteria, and Archaea) of the input sequence using 4 binary numbers. Then, we replicate them, converting input auxiliary sequences into the L×4-dimensional matrix. If there is no provided group information, all entries are designated as 0.

For USPNet, the model architecture comprises two key components: the BiLSTM-attention model and the feature extraction module to predict both signal peptide type and cleavage sites. 
\subsubsection*{BiLSTM module}
We first describe the BiLSTM module. It gets the L×20-dimensional embedding vectors of protein sequences and feeds them into one fully connected layer with 252 hidden units.  The output of the fully connected layer has two usages. First, it is passed to two CNN layers to generate L×512-dimensional representations for future use. The CNN can capture the important motifs and aggregate both useful local and global information across the entire input sequence. Second, the output is concatenated with the auxiliary matrix to form the L×256-dimensional matrix as the input to the BiLSTM layer. It is with 128 hidden units to extract long-distance dependencies of sequences over forward and backward directions. In this way, we integrate information from input sequences at a higher level for later prediction by concatenating the outputs of two directions, which can be written as:
\begin{equation}
\begin{aligned}
\mathbf{h}_{t} &=[\overrightarrow{\mathbf{h}}_{t} ; \overleftarrow{\mathbf{h}}_{t}] \\
&=BiLSTM\left(e^{x^{i}}, \overrightarrow{\mathbf{h}}_{t-1}, \overleftarrow{\mathbf{h}}_{t+1}, \theta_{C l_{s}}\right),
\end{aligned}
\end{equation}
where $\overrightarrow{\mathbf{h}}_{t}$ and $\overleftarrow{\mathbf{h}}_{t}$ denote hidden states at time point t over forward and backward directions, respectively. $e^{x^{i}}$ denotes the embedded representation of the input sequence element $x^{i}$. $\theta_{C l_{s}}$ denotes the parameters of the BiLSTM layer. The BiLSTM layer takes the embedded input $e^{x^{i}}$, the forward hidden state $\overrightarrow{\mathbf{h}}_{t-1}$, the backward hidden state $\overleftarrow{\mathbf{h}}_{t+1}$, and the parameters $\theta_{C l_{s}}$ as input and outputs the concatenated hidden states $\mathbf{h}_{t}$.
The BiLSTM layer outputs 256-dimensional vectors.

To better aggregate information from different feature subspaces, we develop a multi-head attention mechanism in the BiLSTM module, with the scaled dot-product attention in each head:
\begin{equation}
\begin{aligned}
\begin{array}{c}
\text{Attention}(Q, K, V) = \text{softmax}(\frac{QK^{T}}{\sqrt{d_k}})V, \\

\end{array}
\end{aligned}
\end{equation}
where $\frac{1}{\sqrt{d_k}}$ is the scaling factor, and $d_k$ is the dimension of the queries and keys. We conduct Equation (2) in each head and concatenate the multiple single-headed outputs together:   
\begin{equation}
\begin{aligned}
\begin{array}{c}
\text { MultiHead(Q, K, V) }=\text { Concat}\left(\text { head }_{1}, \ldots, \text { head }_{h}\right)W^{O},\\
\text { head}_{i}=\text { Attention }\left(QW_i^{Q}, KW_i^{K}, VW_i^{V}\right)
\end{array}
\end{aligned}
\end{equation}

Where the projections are parameter matrices 
$W_i^{Q}\in \mathbb{R}^{d_{model}\times d_k}$, 
$W_i^{K}\in \mathbb{R}^{d_{model}\times d_k}$, 
$W_i^{V}\in \mathbb{R}^{d_{model}\times d_v}$; Q is the output from the previous BiLSTM layer; both K and V are the same as the input to the BiLSTM layer. 
And h represents the number of heads. We employ $h=2$ to form the parallel attention layers.  For each head, we use $d_{model} = 256$, $d_k=d_v=d_{model}/h=128$. Within the multi-head attention module, we employ a residual connection step to stabilize training and help mitigate the possible vanishing gradient problem. The output of the multi-head attention module is further concatenated with its input, and then element-wise added with the representations of previous CNN layers. After that, the vector is finally fed into the second BiLSTM layer with 256 hidden units. 

To predict the cleavage site, we build a three-layer MLP on top of the second BiLSTM layer. The output of BiLSTM is hence fed into three fully connected layers to produce the L×11-dimensional matrix. The 11-dimensional vectors correspond to SP region labels, with 10 SP region labels for each  amino acid in protein sequences and 1 annotation for vacancies of input sequences with lengths less than 70. The output matrix is used to directly predict the presence of cleavage sites across all positions of amino acids of input protein sequences.

Further, the output of the second BiLSTM layer is also used to predict signal peptide types. The output vector is reshaped and first goes through one fully connected layer to reduce its dimensionality and result in a 512-dimensional vector. 
Meanwhile, we include MSA embeddings here to integrate more information. The pre-trained MSA Transformer model \cite{MSA} generates 768-dimensional embeddings, and we use fully connected layers to aggregate high-level information about these embeddings as a 64-dimensional vector. The processed results are concatenated with the vectors from previous outputs of the fully connected layer. The concatenated results hence have 512+64 channels and are summarized by a fully connected layer with 256 hidden units. Finally, the outputs are forwarded to a normalized linear projection layer to make the prediction.

\subsubsection*{Loss function designed for the data imbalance problem}
Most existing knowledge bases related to signal peptides suffer from extreme data imbalance. The number of the minority classes of signal peptide sequences is usually much smaller than that of non-signal-peptide sequences. This scenario may lead to poor generalization in low-resource data types. Existing techniques, such as cost-sensitive re-sampling and re-weighting, can effectively cope with challenges brought by data imbalance. However, for the signal peptide prediction, these methods are case-sensitive and prone to overfitting. Post-correction methods are also proposed to handle data imbalance, but generalization improvement cannot be ensured. Inspired by the vanilla empirical risk minimization (ERM) algorithm, we introduce LDAM loss and combine it with reweighting to solve the issue.

LDAM loss focuses on correcting the cross-entropy function by introducing the margin item ${\bigtriangleup y}$ to improve the generalization of classes. It suggests that a suitable margin should achieve good trade-offs between the generalization of major classes and minor classes. The class-dependent margin for multi classes is verified to have the form:

\begin{equation}
\begin{aligned}
\bigtriangleup y_{j}=\frac{C}{n_{j}^{1 / 4}},
\end{aligned}
\end{equation}
where $n_{j}$ is the sample size of the j-th class, $C$ is a hyper-parameter to be tuned.

However, the margin item has changed the amplitude of logits of the correct category and influenced generalization, which adds up to the difficulties of representation learning and makes the model sensitive to parameter settings. Here, we introduce agent vectors for each class
by applying a normalized linear layer in the final layer of the classifier \cite{r22}. We normalize both the inputs and the weights of the linear projection layer to clamp the inner product of feature vectors and weight vectors into [-1, 1]. The experiments prove that the normalization can not only improve the robustness of the model but also accelerate the convergence during the training process.

Empirically, softmax cross-entropy loss with scaling factor is widely used in reinforcement learning and relevant fields \cite{liu2018deepnap}. On the one hand, if the scaling factor is too large, class intervals will become close to 0 and therefore influence generalization. On the other hand, a small scaling factor will lead to a deviation from the objective function. Here in our method, a scaling factor is used as a hyper-parameter to ensure the final logits locate in a reasonable scope:

\begin{equation}
\begin{aligned}
\mathcal{L}(x,y)=-\log \left(\frac{e^{\frac{(z_{y}-\bigtriangleup y)}{s}}}{e^{\frac{(z_{y}-\bigtriangleup y)}{s}}+\sum_{i\neq y}^{} e^{\frac{z_{i}}{s}}}\right),
\end{aligned}
\end{equation}
where
\begin{equation}
\begin{aligned}
\bigtriangleup j = \frac{C}{n_{j}^{1 / 4}}\  for\  j \in \{ 1,...,K\},
\end{aligned}
\end{equation}
$ {z_{y}}$ denotes the logit score of the ground truth label, ${z_{i}}$ denotes the logit score of other labels, and s represents the scaling factor.
Both ${z_{y}}$ and ${z_{i}}$ are normalized as agent vectors.

The final objective function optimizes signal peptide prediction and cleavage site prediction jointly:

\begin{equation}
\begin{aligned}
\mathcal{L}_{s}=-\frac{1}{N}\mathlarger{\mathlarger{\sum}}_{j=1}^{N}\mathlarger{\mathlarger{\sum}}_{y=1}^{K}\mathcal{L}(x_j,y_j),
\end{aligned}
\end{equation}
\begin{equation}
\begin{aligned}
\mathcal{L}_{c}=-\frac{1}{N*L}\mathlarger{\mathlarger{\sum}}_{j=1}^{N*L}\mathlarger{\mathlarger{\sum}}_{y=1}^{K'}\mathcal{L}(x_j,y_j),
\end{aligned}
\end{equation}
\begin{equation}
\begin{aligned}
\mathcal{L}_{uspnet}= \mathcal{L}_{s} + \mathcal{\tau}\mathcal{L}_{c},
\end{aligned}
\end{equation}

where $\tau = 1$. $\mathcal{L}_{s}$ denotes the objective function for  signal peptide prediction, where the number of classes $K =6$.  And $\mathcal{L}_{c}$ denotes the objective function for cleavage site prediction, where the number of region labels $K' = 11$.  $N$ is the number of input protein sequences, and $L$ is the maximum length of the input sequence set as 70.

\subsubsection*{Protein language models to enrich representations}
Compared with other protein language models, the MSA transformer \cite{MSA} makes the best of the powerful model architecture and large evolutionary database. In addition, self-supervised learning is good at encoding properties of protein sequences in many different scales, which contributes to the high performance of the model. It is verified that without biological signals other than sequences, the model can still learn the structures of amino acids, protein sequences, and evolutionary homology \cite{r16}.  
Furthermore, the generalization of the model trained over multi-families is better than the effects based on a single family. Accordingly, we believe that by adding the representation of the MSA transformer, the similarity of sequences within the single group can be captured, and the USPNet will become more powerful to differentiate sequences from different group organisms.  

To apply MSA transformer in our method, we first generated Multi-sequence alignment (MSA) for each sequence by searching updated UniClust30\cite{mirdita2017uniclust} with HHblits\cite{steinegger2019hh}. And diversity maximizing subsampling strategy\cite{MSA}, a greedy strategy that starts from the reference and adds the sequence with the highest average hamming distance to the current set of sequences, is used to reduce the number of aligned sequences to 128 for full MSAs with the size larger than 128. Then we input MSAs to the pre-trained MSA Transformer model \cite{MSA} to generate 768-dimensional embeddings from the final layer of the model. 

The other version of our model, USPNet-fast replaces MSA embeddings with ESM-1b embeddings \cite{r16}. ESM-1b takes single sequences as input to generate embeddings, so it enables USPNet-fast to predict signal peptide and cleavage sites faster and without much performance degeneration. 

\subsubsection*{Training details}
In USPNet, we apply the Adam optimizer \cite{adam} with the initial learning rate of 2${\times10^{-3}}$ and a weight decay of 1${\times10^{-3}}$ \cite{r23}. The total number of epochs is set to be 300, with the early stopping strategy to get the best model during training. All the experiments run on four V100 GPU cards with 32GB memory. USPNet does not rely on PSSM and HMM profiles to enrich the embeddings or enhance performances. It is more straightforward to make predictions without evolutionary profile-based features, therefore ensuring a shorter processing time. For SignalP6.0 and SignalP5.0, we used the hyperparameters they provided in the code repositories to train the models.

\subsection{Performance evaluation}
For the taken metrics in the experiments, we summarize them into two components: Metrics used for evaluation of classification performance and cleavage site (CS) prediction performance. For classification, we used the Matthews correlation coefficient (MCC) as a measurement. We considered true/false positives/negatives by labels of sequences. MCC can be written as:

\small
\begin{equation}
\begin{aligned}
\text{MCC}=\frac{\mathrm{TP} \times \mathrm{TN}-\mathrm{FP} \times \mathrm{FN}}{\sqrt{(\mathrm{TP}+\mathrm{FP}) \times(\mathrm{TP}+\mathrm{FN})\times(\mathrm{TN}+\mathrm{FP}) \times(\mathrm{TN}+\mathrm{FN})}},
\end{aligned}
\end{equation}
\normalsize

where TP denotes the number of true positives, TN denotes the number of true negatives, FP denotes the number of false positives, and FN denotes the number of false negatives.
Here, we include both MCC1 and MCC2 to evaluate the performance and hence compute twice. For MCC1, we only take globular and/or transmembrane proteins as negative set and proteins of relevant signal peptide type as positive set. And then, for MCC2, all the rest sequences are added to the negative set.

While in the ablation study, we not only measure classification performance by MCC over different organism groups but also apply overall MCC, Kappa, and balanced accuracy for evaluation. This is because in the ablation study part, some models' performances in minor classes are pretty close, and the overall performances will be more straightforward and intuitive to display differences. Kappa is broadly used in consistency tests and measurements of multi-class classification. And considering the extreme imbalance in datasets, accuracy cannot demonstrate the actual performance of models. Therefore, we take balanced accuracy instead. Balanced accuracy normalizes true positive and true negative predictions over the total number of samples:

\begin{equation}
\begin{aligned}
\text { Balanced accuracy }=\frac{T P R+T N R}{2},
\end{aligned}
\end{equation}
where TPR denotes the true positive rate and TNR denotes the true negative rate.

In the CS prediction part, we take precision and recall as measurements. Precision represents the fraction of correct CS predictions over the total number of predicted CSs, and recall represents the fraction of correct CS predictions over the number of ground truth labels of CSs. Notably, we do not tolerate prediction deviations; in other words, only exact site predictions will be considered correct.

To estimate the quality of multiple sequence alignment, the number of effective sequences (Neff) is applied. It can be calculated as the following function:

\begin{equation}
\begin{aligned}
N_{\text {eff }} = \sum_{i = 1}^{N} \frac{1}{\text { weight }_{i}},
\end{aligned}
\end{equation}
where N is the number of sequences in an MSA, and $weight_{ij}$ is the sequence identity between any two homologous sequences i and j in the MSA.

\subsection{3-D structure prediction}
Considering the computational resource, we apply the ColabFold \cite{mirdita2022colabfold} version of AlphaFold2 to perform the 3-D structure prediction in Figure \ref{agnostic}.d and e. Specifically, we set the number of synchronously running models as 1, and use the amber relaxation to refine the prediction. Other settings remain default. For the visualization of the structure and the following alignment operation, we utilize Pymol \cite{delano2002pymol}. And because the training set has a large number of sequences, to save our time, the structure inferences of the training set and 347 candidates were performed by ESMFold \cite{lin2022language}, and the TM-scores are calculated by TM-align \cite{zhang2005tm} (the subsequent 3 pairs of samples structure alignment was still conducted by ColabFold).

\section{Data availability}
All the datasets we used are listed in the Method part and are publicly available.  All other relevant data supporting the key findings of this study, such as the results of the metagenomics study, are available within the article and the Supplementary Information files or from the corresponding author upon reasonable request. Source data are provided in this paper.

\section{Code availability}
The open source codes of USPNet can be found at https://github.com/ml4bio/USPNet, and the experiments conducted to produce the main results of this article are also stored in this repository. 
  
\newpage
\bibliography{mybib}
\bibliographystyle{myrecomb}


\end{document}